\pdfoutput=1

\documentclass[11pt]{article}

\usepackage{naacl2021}

\usepackage{times}
\usepackage{latexsym}

\usepackage[T1]{fontenc}

\usepackage[utf8]{inputenc}
\usepackage{amsmath}
\usepackage[ruled,linesnumbered]{algorithm2e} 

\usepackage{multicol, blindtext} 
\usepackage{multirow}  
\usepackage[utf8]{inputenc}
\usepackage{graphicx,subcaption}
\usepackage{comment}
\usepackage{latexsym}
\usepackage{soul}

\usepackage{microtype}

\title{ReinforceBug: A Framework to Generate 
Adversarial Textual Examples }

\author{Bushra Sabir \\
  University of Adelaide\\
  CREST - The Centre for \\ Research on Engineering \\Software Technologies   \\
  CSIROs Data61  \\
   \And
  M. Ali Babar \\
  University of Adelaide\\
  CREST - The Centre for \\ Research on Engineering \\Software Technologies   \\
  \\\And
  Raj Gaire \\
  CSIROs Data61  \\}

\begin{document}

\maketitle

\begin{abstract}
Adversarial Examples (AEs) 
generated by perturbing original training examples 
are useful in improving the robustness of Deep Learning (DL) based models.
Most prior works, generate AEs that are 
either unconscionable 
due to lexical errors
or semantically or functionally deviant from original examples. 
In this paper, we present \emph{ReinforceBug},
a reinforcement learning framework, 
that learns a policy that is transferable on unseen datasets and generates utility-preserving and transferable (on other models) AEs.
Our results show that our method is on average 10\% more successful as compared to the state-of-the-art attack TextFooler.
Moreover, the target models have on average 73.64\%
confidence in wrong prediction, the generated AEs preserve the functional equivalence and semantic similarity (83.38\% ) to their original counterparts, and are transferable on other models with an average success rate of 46\%.
\end{abstract}

\section{Introduction}
Machine Learning (ML) models have attained remarkable success in several tasks 
such as classification and decision analytics. 
However, ML Models specifically Deep Learning (DL) based models, are often sensitive to Adversarial Examples (AEs).
AEs consist of transformed original training data samples 
that preserve the intrinsic utilities of the ML solutions, 
but influence target classifier’s predictions between the original and the modified input \citep{fass2019hidenoseek}.
Recent works \citep{fass2019hidenoseek,li2020bert,biggio2018wildpatterns} have demonstrated that 
(i) including AEs as a part of training data can enhance the robustness and generalization of the ML models, and 
(ii) these examples can be utilized to test the robustness of current ML models and help in understanding their security vulnerabilities and limitations.
Previous works on generating AEs have attained success in 
image \citep{biggio2018wildpatterns} and on few conventional text classification tasks such as sentiment, text entailment and movie reviews \citep{li2018textbugger,jin2019bert,li2020bert}. 
Nevertheless, generating AEs for discrete textual data is still a challenge~\citep{jin2019bert}.
Textual AEs generated by the prior works have the following limitations:
%
\paragraph{Utility Preservation}
Textual AEs require to satisfy task-specific constraints 
such as lexical rules (spellings or grammar), 
semantic similarity 
and functional equivalence to original examples.
Yet, most of the current state-of-the-art methods do not satisfy these constraints, 
thereby generating imperceivable AEs for end-users 
\citep{zhang2020adversarial}.
A few recent works \citep{li2020bert,wang2019natural,li2018textbugger,jin2019bert} have considered semantic similarity constraint, 
but other constraints have been barely explored \citep{jin2019bert}.
 
\paragraph{Knowledge Transferability}
Most prior works \citep{alzantot2018generating,jin2019bert,wang2019natural}  generate one to one example-specific AEs, i.e., for a given example $x$ they generate an example $x'$. 
Each example $x$ is considered independent in a corpus $C$, and no relationship between different examples in the corpus is assumed.
Therefore, the knowledge gained by transforming an example to AE is limited to a single example and is not reused on other examples.  
This process is both time-consuming and may not generalize the identified vulnerabilities of the target model.

\paragraph{Word Replacement Strategy}
 Most prior works use a single word replacement strategy such as synonym substitution \citep{alzantot2018generating,jin2019bert,wang2019natural} or character perturbation \citep{gao2018black} to generate AEs. 
This strategy has two main disadvantages: (i) 
the AEs generated using a character-based replacement strategy,
contain many spelling errors that results in unnatural texts; and
 (ii) multiple word transformation with a synonym in a synonym-based replacement strategy affects the fluency of the language, making it sound unnatural.
 For instance \citep{jin2019bert} generated AE 
``Jimmy Wales is a big, 
\st{fucking idiot liar} \textcolor{red}{friggin nincompoop deception}" which sounds unnatural and is grammatically incorrect. 

\paragraph{Handling Noisy Datasets}
Most prior works generated AEs for 
datasets such as Yelp \citep{YelpData30:online} and Fake News \cite{FakeNews89:online} 
which do not contain many spelling mistakes or out-of-the-vocabulary words.
However, human generated natural text is prone to lexical errors.
For example, 
tweets usually contain informal language, misspellings and unknown words. 
Prior works, such as \citep{jin2019bert, wang2019natural,alzantot2018generating, li2020bert} that 
considered synonym substitution 
strategy,
cannot deal with such noisy dataset. 

\subsection*{Our contribution}
We present \emph{ReinforceBug}
that addresses the aforementioned limitations of prior works. Our code is available at \url{http://shorturl.at/dkmuP}.
%
Our main contributions are summarized as follows:
\begin{enumerate}
\item  We propose a reinforcement learning framework, 
\emph{ReinforceBug}, that 
learns a policy to generate utility-preserving AEs
under the black-box setting 
with no knowledge about the target model architecture or parameters. 
\item We evaluate \emph{ReinforceBug} on four state-of-the-art deep learning models 
Email spam, Twitter spam, Toxic content and  Review polarity detection respectively.
\item We investigate the transferability of our learned policy on a new dataset. 
\item We also examine the transferability of our generated AEs on three other state-of-the-art deep learning models. 
\end{enumerate}
\noindent
\begin{figure*}[!tb]
\centerline{\includegraphics[width=0.9\textwidth]{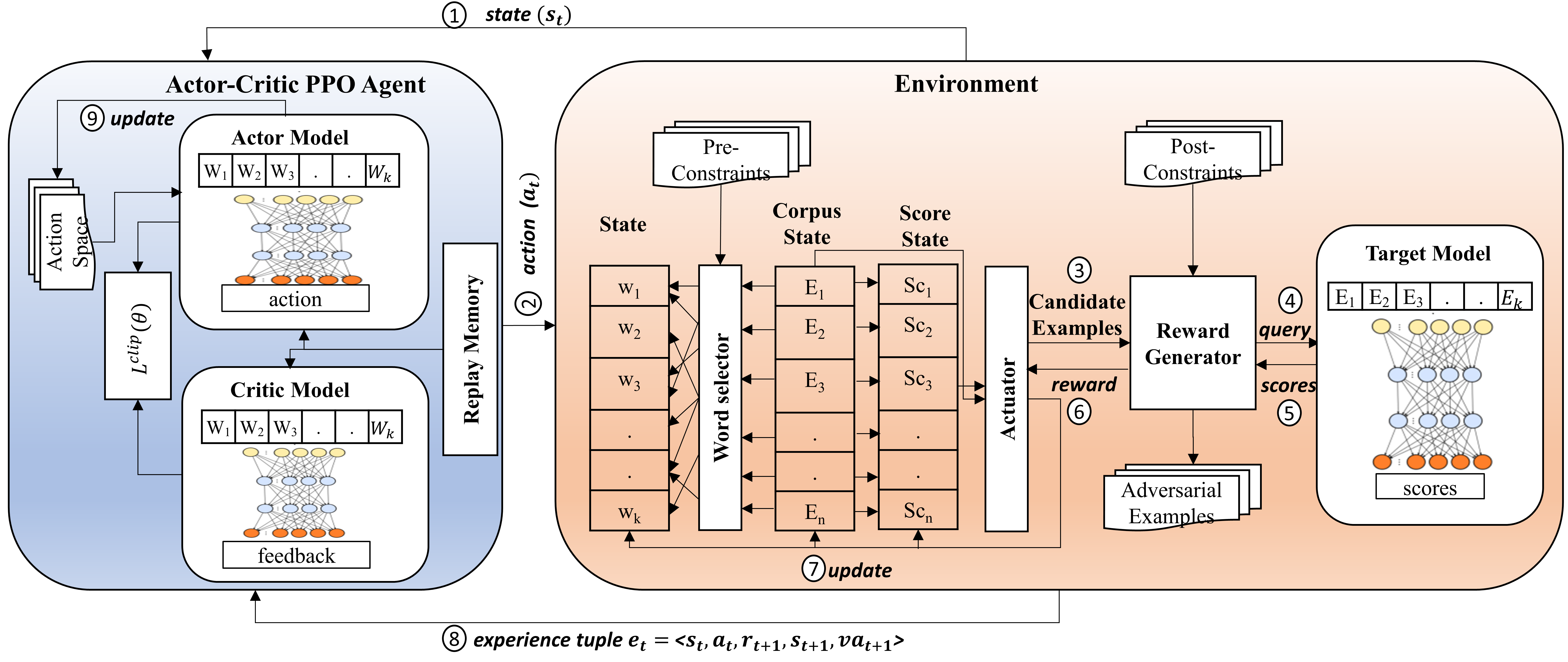}}
\caption{Overall of ReinforceBug}
\label{figure0}
\end{figure*}
\section{Related Works}
\label{relatedworks}
Adversarial attacks are extensively studied in computer vision domain \citep{biggio2018wildpatterns,goodfellow2014explaining}. 
Early works in adversarial text attacks were inspired by Generative Adversarial Networks (GANs) \citep{wong2017dancin,zhao2018generating}.
\citep{wong2017dancin} showed that GAN-based reinforcement learning algorithms become unstable after a few perturbations. 
Later, heuristic-based methods such as word removal \citep{ebrahimi-etal-2018-hotflip}, Out-Of-Vocabulary (OOV) words  \citep{gao2018black} and synonym replacement \citep{li2018textbugger,jin2019bert,alzantot2018generating} have been proposed.
Among these studies, \emph{DeepWordBug} \citep{gao2018black} generates AEs by randomly transforming a word by OOV word in an example. 
This approach is practical in producing AEs efficiently; however, it generates AEs that can be detected by the end-user due to large proportion of lexical errors.
\emph{TextBugger} \citep{li2018textbugger} proposed an attack framework to generate adversarial samples using the multi-level approach. 
It identified important words for each example and replaced them with optimal bugs.
A recent attack, \emph{TextFooler} \citep{jin2019bert} generates utility-preserving AEs by replacing an important word in an example with its synonym and 
considers grammatical equivalence and semantic similarity constraint.
We compare our method with \emph{TextBugger} and \emph{TextFooler}.

\section{ReinforceBug}
\label{reinforcebug}
\subsection{Definitions}
\label{definitions}
\paragraph{Prediction Confidence Score (PCS)}
\label{PCS}
The PCS($x$,$y$) of model $F$ depicts the likelihood of an input $x \in X$ having a label $y \in Y$.
The smaller PCS($x$,$y$) suggests $F$ has low confidence that $x$ has a label $y$.
\paragraph{Utility-Preserving AEs} 
\label{AE}
Given a real example $x$ having a label $y$,
an AE against $x$ is an input $x'$ $\leftarrow$ $x+\imath$ with a minor perturbation $\imath$.
$x'$ is lexically, semantically and functionally similar to $x$ but $F$ predicts an incorrect label for it with high PCS 
i.e., $y' \leftarrow F(x')$  such that $y' \neq y$ and PCS($x'$,$y'$)$>\nu$. 

\subsection{Problem Formulation} 
Given a pre-trained target model $F$, 
we need to 
simulate a non-targeted black-box attack \citep{morris2020textattack} to generate a set of utility-preserving AEs
$A_{\mathrm{exp}}$ from a Corpus $C$ with $N$ examples and having corresponding target label $t_g \in Y$. 
Furthermore, it should 
learn a policy $\pi_{\theta(s, a)}$ to perform perturbation $\imath$ on $C$ such that 
the model generates AEs with high PCS and 
semantic similarity with the original example but have low perturbation rate and lexical errors.
Moreover, the policy should be transferable on unseen datasets.
\subsection{System Overview}
Fig~\ref{figure0} provides an overview of ReinfoceBug.
We model an attack as a reinforcement learning  \citep{sutton2018reinforcement} process consisting of three main components: 
an environment, 
Proximal Policy Optimization (PPO) agent \citep{schulman2017proximal} and
action space.
Firstly, the environment state ($s_t$) at time $t$ is processed as input by an agent followed by an action $a_t \in A$ determined by an agent to update $s_t$ to the next state $s_{\mathrm{t+1}}$, 
where $A$ represents an action space (set of valid actions given the state).
Subsequently, the environment's actuator acts on the corpus state to construct candidate examples. 
These examples are then sent to the reward generator module, 
which is responsible for computing the reward $r_t+1$ for action $a_t$.
The reward generator applies post constraints $Post \subset P_{\mathrm{const}}$, queries the target model $F$ and obtain scores of the candidate examples to calculate the reward of action $a_t$. 
The reward is sent back to the actuator which determines a valid update to the corpus state as well as the next environmental state $s_{\mathrm{t+1}}$. 
The experience consisting of $<s_t,a_t,r_{\mathrm{t+1}},s_{\mathrm{t+1}}, va_{t+1}>$ is sent to the agent and the agent model is updated. 
Here $va_{t+1}$ shows the valid actions mask for next state. 
The $va_{t+1}$ is then used by the agent to update an action space $A$ and restrict the agent to only select valid action on the new state $s_{t+1}$. Each of the modules is discussed below:
\subsubsection{Agent}
We use a customized version of Proximal Proximate Optimal (PPO) Reinforcement Learning (RL) agent with action mask capability \citep{tang2020implementing}. 
PPO is an enhanced version of Actor-Critic \citep{grondman2012survey} (AC) Model. 
In AC architecture the agent consists of a separate policy and Q-value network. 
They both take in environment state $s_t$ at each time step as an input, 
while actor determines an action $a_t$ among the possible valid action set $\alpha$ and critic yields value estimation $V_t(s_t)$. 
While an actor uses a gradient descent algorithm to learn a policy that maximizes the rewards $R_t$, 
the critic learns to estimate $R_t$ via minimizing a temporal loss. 
Further, the PPO algorithm avoids large and inefficient policy updates by constraining the new policy updates to be as close to the original policy as possible.
We have selected this agent because our action and state space is substantially large and to avoid enormous policy updates which make the agent unstable (please refer to \citet{schulman2017proximal} for more details).
\subsubsection{Environment}
The environment takes action $a_t$ from the agent as input and outputs the experience tuple $e_t$, AEs $A_{\mathrm{exp}}$ and a flag $done$ depicting the success of the agent in achieving the goal.
\subsubsection{States}
The environment maintains following states.

\emph{Corpus State $C_t$}: 
it is given by $C_t={E_1,E_2,....,E_N}$,
where $N$ is the number of examples in the corpus and $E_i$ is the set of words $W_i={w_1,w_2,...,w_n}$ of an example $i$ at time $t$.

\emph{Score State $score_t$}: 
represents the PCS of target model 
that $C_t$ has the ground truth label, 
i.e., for an example $E_i$ the score of $i$ at time $t$ is given by $score_t[i]= PCS(E_i,tg)$ where $E_i \in tg$ class.

\emph{Environment State $s_t$}
This state $s_t={w_1,w_2,...,w_k}$ is observable by an agent. 
It consists of $k$ mutable important words in corpus $C$.

\emph{Success rate $success_t$}: 
it is the proportion of utility-preserving AEs compared to all AEs generated by the agent at time $t$.
    

\subsubsection{Components}
The environment contains following components:
\begin{table}[!tb]
\caption{List of Actions}
\label{actions}
\centering
\resizebox{\columnwidth}{!}{\begin{tabular}{l|l|l}
\hline
\textbf{Action} & \textbf{Description} &\textbf{ Example}  \\
\hline
Homoglyph & Replace a char with visually similar char.& potentially vs potentia1Iy \\ 
\cline{2-3}
Insertion & Insert a char in a word. & nearly vs n3early \\
\cline{2-3}
BitSquatting&Replace a character with one bit & clearly vs glearly \\
&different char.&\\
\cline{2-3}
Omission&omit one char. & standard vs stanard\\
\cline{2-3}
Addition&Add a char on start or end of a word. & classified vs classifiedf\\
\cline{2-3}
Repetition& Repeat the previous char in the word.& requested vs rrequested\\
\cline{2-3}
get\_synonyms& Replace the word with its synonym&  control vs dominance\\
\cline{2-3}
get\_semantic& Append the word with a word contextually&   like vs such like\\
&related word&\\
\cline{2-3}
get\_syntactic&Replace the word with a word having similar &  bta vs bat\\
&syntax.&\\
\cline{2-3}
currency\_word& If word contains currency symbol replace it & \$ vs dollar\\
&with currency word and vice-versa. & dollar vs \$\\
\cline{2-3}
word\_to\_num&If word represents a number replace it with  & eigth vs 8	\\
&the number.&\\
\cline{2-3}
num\_to\_word& If word is a number replace it to English word. & 8 vs eight\\
\cline{2-3}
word\_weekday& If word is a weekday, replace it with its & Wednesday vs wed \\
&abbreviation or vice-versa&wed vs Wednesday\\
\cline{2-3}
word\_month& If word is a month, replace it with its & August vs Aug \\
& abbreviation vice-versa&aug vs August\\
\hline
\end{tabular}}
\end{table}
\paragraph{Word selector} This component takes $C$ state at $t=0$ and pre-constraints $Pre \subset P_{\mathrm{const}}$ as input.
$Pre$ are task-specific perturbation restrictions on the specific entities in $E_i$. 
For example, spam messages mostly contain URLs, IP addresses, organization names and email addresses pointing to phishing websites.
Perturbing these entities in the spam message can change the functional semantics of the message \citep{ranganayakulu2013detecting}.
Hence, imposing pre-constraints ensures that functional equivalence of $E_i$ after applying perturbations.
To achieve this, we have designed a \textit{countvectorizer} \citep{sklearnf17:online} using a customized tokenizer.
The tokenizer finds the list of immutable entities such as URLs and IP addresses in the text using regular expressions or named entity models \citep{florian2003named}. 
After that, these words are segmented into immutable words $Im_{words}$ using word tokenizer and saved for each example to be utilized later by actuator module. 
For training \emph{ReinforceBug}, our method first computes the important words from the training datasets as state and then learns a policy to identify best actions to transform the state to a new state such that the success of the attack is maximum.
The important words are selected using a word importance score $I_{w_{idx}}$. 
The $I_{w_{idx}}$ is calculated as the sum of prediction change in all the examples (k) containing $w_{idx}$ before and after deleting $w_{idx}$. The candidate words $w_{idx} \in$ most frequent words in the training dataset vocabulary. 
It is formally defined as follow.
\begin{equation}
I_{w_{idx}}= 
\begin{cases}
\frac{
   \sum\limits_{i=1}^{k}(
   \frac{|F(E_i)- F(w_{idx} \notin E_i)|}{F(E_i)}
   )}{k}
 \end{cases}
 \end{equation}
If the $I_{w_{idx}}$ is $>0$, we consider it as an important word. The final list of all the important word is considered as the state $s{\mathrm{t}}$. 
 $s{\mathrm{t}}$ and mapping $C_{\mathrm{map}}$ that maps each word to the corpus examples is sent back to the environment. For testing, the designed \textit{countvectorizer} is used to transform the testing data onto these selected words.
\paragraph{Actuator} 
This module is responsible to execute an action $a_t$ selected by an agent. Firstly, the actuator transforms action $a_t$ s into an action tuple $<w_{\mathrm{idx}},act_{\mathrm{idx}},rep_{\mathrm{idx}}>$, where $w_{\mathrm{idx}}, act_{\mathrm{idx}}$ and $rep_{\mathrm{idx}}$ depict 
the index of the word to be replaced, 
the operation to be performed on that word and 
replacement word index respectively. 
Subsequently, example indexes $E_{\mathrm{idx}}$ containing the word $w_{\mathrm{idx}}$ are obtained by querying the $C_{\mathrm{map}}$.
After that, the actuator examines the score state $score_t[k]$ of each example $k$ in $E_{\mathrm{idx}}$.
If $score_t[k]> \nu$, only then it is selected as an example to be perturbed, here $\nu$ represent the PCS threshold.
In this way, the examples for which AEs have already been found are not perturbed further and other examples are given a chance.
Once these examples are selected, the operation $act_{\mathrm{idx}}$ is applied to $w_{\mathrm{idx}}$ which results in multiple replacement options. 

Table~\ref{actions} provides the list of operations considered.
For example, if the action is Homoglyph and the word indexed by $w_{\mathrm{idx}}$ is ``solid" then following replacements can be done "so1id,sol1d,s0lid,5olid".
The new word $w_{\mathrm{new}}$ is selected by $rep_{\mathrm{idx}}$.
After $w_{\mathrm{new}}$ is selected against the $w_{\mathrm{old}}$, 
if $w_{\mathrm{old}}$ in not in the immutable word list of example k than a candidate AEs are generated by substituting $w_{\mathrm{old}}$ with $w_{\mathrm{new}}$ in the previously selected examples.
In this way the functional equivalence is ensured before generation of AE.
The selected candidate examples are then sent to the reward generator.
The reward generator module returns the reward of changing the $w_{\mathrm{idx}}$ in the selected examples by applying $act_{\mathrm{idx}}$ and selecting $rep_{\mathrm{idx}}$.
The reward generator also outputs the AEs that satisfies all the post constraints.
Finally, the state is updated by replacing the $w_{\mathrm{idx}}$ in $s_t$ with index of new word $w_{\mathrm{new}}$.
All further actions on $w_{\mathrm{idx}}$ are then invalidated. 
This is done by setting the $va_{\mathrm{t+1}}$ for all actions on the $w_{\mathrm{idx}}$ in an action\_space to False.
In this way, multiple actions on the same word cannot be performed in one training episode. 
The $success_{t+1}$ is updated, the episode completes if the $success_{t+1}$ for the corpus state $C_{\mathrm{t+1}}$ has reached the threshold specified by $\phi$ or all the words in the state have been updated.
The module constructs an experience tuple $e_t$ for the agent and returns $e_t$, $C_{\mathrm{t+1}}$, $score_{\mathrm{t+1}}$ and $success_{\mathrm{t+1}}$ to the environment.
\paragraph{Reward Generator (RG)} 
\begin{algorithm}[!tb]
   \caption{Reward\_Generator}
   \label{reward}
   \begin{small}
   \SetKwProg{generate}{Function \emph{generate}}{}{end}
   {\bfseries Input:} 
   Original examples $org$,
   candidate examples $cand$,
   time $t$,
   $Post \subset P_{\mathrm{const}}
   $\;
   {\bfseries Output:} 
   $
   Adv_{\mathrm{exp}}
   $,
   reward 
   $r_{t+1}$,
   $C_{\mathrm{t+1}}$
   \;
 \textit{ \textbf{Initialize reward} $\leftarrow 0$,  $Adv_{\mathrm{exp}}$ $\leftarrow \{\}$}\;
      updated\_cand, {rewards}\;

     \eIf{($t=0$)} 
     {
     $
     updated\_cand=cand
     $\;
     \textit{\textbf{Query Target Model}}\;
     $
     score_{\mathrm{t}} \leftarrow target\_model.query(cand)
     $\;
     }
     {
     \textbf{Initialize thresholds for \textit{spellerror} $\eta$,  \textit{gramerror} $\iota$, \textit{Semantic}  $\varepsilon \in$ Post}\;
      $N' \leftarrow $ total candidate examples\;
      \ForAll{$c_k \in cand$}{
      $sem=
      Semantic(org_k,c_k)
      $\;
      $
      spell=\frac{spellerror(c_k)}{wordlen}
      $\;
      $
      gram=\frac{gramerror(c_k)}{wordlen}
      $
      \;
     \If
     {
     $
     sem>\varepsilon
     $ 
     \textbf{and} 
     $ spell 
     <=\eta
     $
     \textbf{and}
     $ 
     grammar <= \iota)
     $
     }
     {
    \textit{ \textbf{ Query target\_model}}\;
         $new\_sc_k=query(c_k)$
         $prev\_sc_k \leftarrow score_t[k]
         $\;
         $change[k] =\frac{prev\_sc_k- new\_sc_k}{prev\_sc_k}
         $\;
     
    \If{
        $change[k] > 0
        $
        }
        {
         \textit{\textbf{update the corpus state}}\;
         $
         C_{\mathrm{t+1}} \leftarrow$
         $ E_{\mathrm{idx}}[k] \leftarrow c_k
         $\;
         \textbf{update the score state}\;
         $
         score_{\mathrm{t+1}}[k] \leftarrow new\_sc_k
         $\;
         }
        \textbf{ Compute Reward }\;
        \uIf{($
        score_{\mathrm{t}}[k] < \nu
        $)}
         {
         $
         Adv_{\mathrm{exp}}.append(c_k)
         $\;
         $
         p_{rate}[k]=\frac{words perturbed}{wordlen}
         $\;
          $
           r_{t+1}+=\frac{\frac{(org_{score[k]}-score[k])}{org_{score[k]}}+sem[k]}{spell[k]+gram[k]+p_{rate}[k]}
          $
          \
         }
         \Else
         {
         $
         r_{t+1}+= change[k]
         $ \;
         }
       }}
        $r_{t+1}$=$\frac{r_{t+1}}{N'} $\; 
}    
 {\bfseries return:}
 $
 Adv_{\mathrm{exp}},r_{t+1}, C_{\mathrm{t+1}}
 $\;
 
\end{small}
\end{algorithm}

Algorithm \ref{reward} shows the pseudo-code of the RG module.
RG takes candidate examples, the $C_t$, time $t$ and post constraints as input and outputs the $Adv_{\mathrm{exp}}$, the reward $r_{\mathrm{t+1}}$ of $a_t$ and updated corpus state $C_{t+1}$.
In this study, we have considered three main post utilities that the generated $Adv_{\mathrm{exp}}$ should preserve apart from changing the output of the classifier. 
Firstly, the percentage of the spelling and grammatical errors should not be more than $\eta$ and $\iota$ respectively, and the semantic similarity between the original and AE should be $>\varepsilon$.
If the candidate example meets all these utilities, then the target\_model is queried to obtain the score of the candidate example. 
RG updates the corpus and score states for an example where the difference between previous (before perturbation) and new (after perturbation) score of an example $>0$ . 

Subsequently, the reward generator checks that if the candidate example score has converged to $< \nu$. 
In that case, the example is added to the list of valid AE against the original example, and a reward $r_{t+1}$ is computed as shown in line 30. Otherwise, the sum of the change in the scores of an example is added as reward, as shown in line 31.
In line 30, the  $r_{t+1}$ represents the summation of reward attained on successfully transforming the original examples into well-constrained AE. 
It is calculated as a summation of the change in original and current score of the example k, assisted by semantic similarity with the original example and penalized by lexical errors (spelling and grammatical) and perturbation rate $p_{rate}$ for generating each AE. 
In this way, the agent learns to generate AEs with minimum perturbations and lexical errors and more PCS and semantic similarity. Lastly, the $r_{t+1}$ is normalized by the factor N,' so that the agent can learn the impact of action $a_t$ on a single example irrespective of the size of the corpus.
\paragraph{Target Model}
We have considered black-box access to the target model.
The target model can be any Deep learning model that provides $PCS$ (section \ref{PCS}) score as output .

\section{Experimental Setup}
\label{experiments}
This section presents our experiment details.
\subsection{Datasets}
We study the effectiveness of \emph{ReinforceBug} on three noisy and one conventional text
classification tasks respectively.
Table \ref{dataset} enlists the statistics of the dataset.
Precisely during target model training, we held out 30\% of the training set as a validation set, and all parameters are tuned based on it.
After that, the testing dataset was used to evaluate the performance of the model.
For training and testing our \emph{ReinforceBug}, we split the testing dataset into training (70\%) and testing dataset (30\%). 
The stratified split is applied to ensure the class distribution on these datasets remains consistent with the actual testing dataset.

\subsection{Attacking Target Models}
For each dataset, we train four state-of-the-art models namely Word Convolution Neural Network (CNN) \citep{jain2018spam}, Character CNN \citep{zhang2015character}, 
Word Bidirectional Long Term Short Memory (BiLSTM) \citep{zhou2016text} and Recurrent CNN \citep{lai2015recurrent} on the training set.
We used the same parameter for training these models as provided by the open-source GitHub repository \citep{dongjunL40}.
Table~\ref{tab2} shows the performance of each model on the testing set.
From these models, we selected the models with best performance accuracy as target models (as highlighted in bold) to train \emph{ReinforceBug} . 
For the rest of the models trained on a similar dataset, we studied the transferability of our generated AEs.
Moreover, we test our \emph{ReinforceBug} against an unseen dataset to study the transferability of attack on other datasets.
Lastly, for get\_semantic and get\_synonym action types \citep{pennington2014glove} and \citep{mrksic:2016:naacl} embeddings have been used.
\begin{table}[!b]
\caption{The Target Models training and testing data sets statistics }
\label{dataset}
\centering
\begin{small}
\resizebox{\columnwidth}{!}{\begin{tabular}{c|c|c|c|c|c}
\hline
     \multirow{2}{*}{\textbf{Dataset}} & \textbf{Training} & \textbf{Testing} & \textbf{Avg } &\textbf{Avg Spell-} &\textbf{Avg Gram-}\\
     &\textbf{Data}  &\textbf{Data}  & \textbf{Length} &\textbf{ing errors} &\textbf{mar errors }\\\hline
	 Enron \citep{SpamDatasets} & 28.6k & 9.9k & 244&3.07\%
&	31\%\\ 
     \hline
	 Twitter \cite{Twitter-spam:online} & 8.1k & 3.9k & 15&10.67\%&28.46\%\\ \hline
	Toxic \cite{ToxicCom95:online} & 159.5k & 12.4k  &70&2.13\%&25.60\%\\ \hline
    Yelp \citep{YelpData30:online}& 560k & 38k &139 &0.81\%&21.67\%\\ \hline
\end{tabular}}
\end{small}
\end{table}
\begin{table}[!b]
\caption{Balanced Accuracy of target models on test datasets}
\label{tab2}
\centering
\begin{small}
\resizebox{\columnwidth}{!}{\begin{tabular}{l|c|c|c|c}
\hline
\textbf{Dataset} &
\textbf{WordCNN}&
\textbf{CharCNN}&
\textbf{BiLSTM}&
\textbf{RCNN} 
\\
\hline 
Enron &97.50\%&96.50\%&97.60\%&\textbf{98.30}\%\\
\hline
Twitter&
93.44\%&
92.02\%&
93.72\%&
\textbf{94.15}\%\\
\hline
Toxic&69.05\%&
86.62\%&
\textbf{89.61}\%&
88.56\%\\
\hline
Yelp&94.74\%&
94.44\%&
\textbf{95.60}\%&95.34\%
\\
\hline
\end{tabular}}
\end{small}
\end{table}
\subsection{PPO Agent}
We used stable-baselines \citep{stable-baselines} reinforcement learning library to implement our PPO agent.
We used a Multilayer perception (MLP) model, as an actor and critic models. 
For training the agent, we used 30 episodes for each model. 
\subsection{Utility-Preservation Constraints}
To ensure the functional equivalence,
we defined immutable tokens as pre-constraints using Named Entity Model provided by spacy \citep{Linguist48:online} and regular expressions. 
For Enron dataset, names (Person or Organization), IP, email addresses and URL, while for Twitter and Toxic datasets URL and @Reference and lastly, for Yelp dataset, names and URL are considered as immutable entities.
For semantic and lexical equivalence, we defined three post constraints, i.e., the semantic similarity, which is calculated using Universal Sentence Encoder (USE) \citep{cer2018universal}, we considered $\varepsilon=0.60$.
Moreover, for counting spelling mistakes \citep{symspell:online} is used while for calculating grammar issues, we used language tool \citep{language27:online}. 
\begin{table*}[!htb]
\caption{Samples of Examples (red font depicts the change in the original example)}
\label{examples}
\centering
\begin{small}
\resizebox{\textwidth}{!}{\begin{tabular}{  c | c |c | c | c | c }
\hline
	\textbf{Task: Twitter}  & \textbf{Original Label}  & \textbf{Adversarial Label } & \textbf{Task:  Toxic/Non- }& \textbf{Original Label}  & \textbf{Adversarial Label}\\
	Spam/Benign&Benign (PCS 100\%)&Spam (PCS 98\%)&Toxic Comment&Toxic (PCS 97\%)&non-Toxic (PCS 83\%)\\\hline
	\multicolumn{3}{c|}{I should of just gone home yesterday and spent my day off today with my family \textcolor{red}{relatives}}&  \multicolumn{3}{c}{Even trolls \textcolor{red}{surely} deserve to eat, bastard\textcolor{red}{x}.} \\
    \hline
	\textbf{Task: Email}  & \textbf{Original Label} & \textbf{Adversarial Label}  & \textbf{Task: Positive/}& \textbf{Original Label }& \textbf{Adversarial Label}  \\ 
	Spam/Benign&Spam (PCS 100\%)&Benign (PCS 96\%)&Negative Review &Negative (72\%) &Positive (PCS 86\%)\\\hline
	\multicolumn{3}{c|}{Subject:soul mate one of your buddies hooked you up on a date \textcolor{red}{scheduled} with another buddy.}&\multicolumn{3}{c}{Premium price for a standard sandwich. Would have been an \textcolor{red}{amazingly}  } \\ 
	\multicolumn{3}{c|}{ your invitation:a free dating web site created by women no more invitation:}&\multicolumn{3}{c}{enjoyable visit had the deli guy been at least welcoming .}\\
	\hline
\end{tabular}}
\end{small}
\end{table*}
\begin{table*}[!htb]
\caption{ReinforceBug Attack on Training Dataset}
\label{tab3}
\centering
\begin{small}
\resizebox{\textwidth}{!}{\begin{tabular}{c|l|c|c|c|c|c|c|c}
\textbf{Dataset}&\textbf{Attacks}&\textbf{Success}&	\textbf{Perturbation}&	\textbf{\% of URLs }	&\textbf{Avg PCS}&	\textbf{Avg Semantic}	&\textbf{Avg Spelling}	&\textbf{Avg Grammar}\\
&	&\textbf{ Rate}&	\textbf{Rate}&	\textbf{Perturbed}	&&	\textbf{ Similarity}	&\textbf{Errors}	&\textbf{Errors}\\
\hline
\multirow{3}{*}{\textbf{Enron RCNN}}
    &TextBugger&	\textbf{43.09\%}&	46.64\%&	35.21\%&	62.52\%&	61.51\%&	13.39\%&	35.7\%\\
	&TextFooler&	22.31\%&	\textbf{10.60\%}&	9.09\%&	59.76\%&	76.62\%&	\textbf{3.08\%}&	21.80\%\\
	&\textbf{ReinforceBug}&33.21\%&	11.09\%&	\textbf{0.00\%}&	\textbf{73.82\%}&	\textbf{85.66\%}&	3.30\%&\textbf{	9.88\%}\\
\hline
\multirow{3}{*}{\textbf{Twitter RCNN}}
    &TextBugger&	\textbf{65.98\%}&	24.74\%&	97.67\%&	74.6\%&	58.8\%&	25.77\%&	35.64\%\\
	&TextFooler&	12.66\%&	11.89\%&	99.10\%&	79.17\%&	73.79\%&	5.22\%&	15.68\%\\
	&\textbf{ReinforceBug}&14.99\%&	\textbf{11.29}\%	&\textbf{0.00}\%&	\textbf{82.82}\%&\textbf{80.86}\%&\textbf{4.14}\%&	\textbf{13.99}\\
\hline
\multirow{3}{*}{\textbf{Yelp BiLSTM}}
    &TextBugger&	\textbf{44.88\%}&	21.89\%&	6.25\%&	58.62\%&	66.97\%&	5.22\%&	16.85\%\\
	&TextFooler&	33.80\%&	\textbf{4.15\%}&	4.73\%&	58.39\%&	80.60\%&	\textbf{0.74\%}&	14.13\%\\
	&\textbf{ReinforceBug}& 39.35\%&	\textbf{3.18\%}&	0.00\%&	\textbf{63.78\%}&\textbf{87.78\%}&	1.88\%&	\textbf{9.68\%}\\
\hline
\multirow{3}{*}{\textbf{Toxic BiLSTM}}
    &TextBugger&	\textbf{42.11\%}&	24.78\%&	45.10\%&	62.85\%&	60.83\%&	11.36\%&	19.98\%\\
	&TextFooler&	21.00\%&	\textbf{7.07\%}&	28.57\%&	64.17\%&	76.68\%&	\textbf{1.53\%}&	14.70\%\\
	&\textbf{ReinforceBug} &31.22\%	&9.43\%	&\textbf{0.00\%}&	\textbf{71.52\%}&	\textbf{78.07\%}&	3.69\%	&\textbf{6.40\%}\\
\hline

\end{tabular}}
\end{small}
\end{table*}
\section{Results}
\label{results}
Results of our experiments are presented here.
\subsection{Attack Evaluation}
\label{evaluate}
Table \ref{examples} shows the samples of AEs generated by  \emph{ReinforceBug} and Table~\ref{tab3} illustrates the main results
our experiments.
It can be seen that \emph{ReinforceBug} produces AEs with a comparatively high PCS (i.e., on average 74\%),  semantic similarity (i.e., on average 83.5\%) than other two attacks for all the dataset. 
However, its success rate 
is on average 15\% less than TextBugger \citep{li2018textbugger} for the all the models.
One reason behind it is that TextBugger generates unrealistic AEs with least PCS (i.e., on average 66\%), low semantical similarity (i.e., on average 63\%) 
and significantly high lexical errors (i.e., on average 14\% spelling and 27\% grammar errors).
Additionally, TextBugger on average perturbs more than 59\% word of the original text to generate AEs which is too large to be ignored by the end-user.
Moreover, Enron and Twitter datasets, 
TextBugger 
perturbs 9.09\% and 97.67\% of the URLs present in the text, thus adversely effecting the functional semantics of the text.
Therefore, the success rate of TextBugger is an overestimation and deviates from the semantic, functional and lexical constraints.

In comparison with TextFooler \citep{jin2019bert} our method has significantly high success rate (i.e., on average more than 10\%) for all the models. 
It was expected because TextFooler relies on synonym substitutes technique to generate AE, however, for noisy datasets with relative high lexical errors such as Twitter (Table \ref{dataset}) this method tends to fail. 
Lastly, TextFooler produces 5\% more grammatical errors than \emph{ReinforceBug} and similar to TextBugger, TextFooler also perturbs URLs, thus effecting the functional semantics of the generated AE.
These findings suggest that \emph{ReinforceBug} produces effective and utility-preserving AEs.
\begin{table*}[!htb]
\caption{ReinforceBug Attack on Testing Dataset (Test time is given days-hours:minutes:seconds format)}
\label{testing}
\centering
\begin{small}
\resizebox{\textwidth}{!}{\begin{tabular}{c|l|c|c|c|c|c|c|c}
\textbf{Dataset} &\textbf{Attacks}&\textbf{Test}&\textbf{Success}&	\textbf{Perturbation}&\textbf{Avg PCS}&	\textbf{Avg Semantic}&\textbf{Avg Spelling}&\textbf{Avg Grammar}\\
&&\textbf{Time}&\textbf{ Rate}&\textbf{ Rate}&&	\textbf{ Similarity}	&\textbf{Errors}	&\textbf{Errors}\\
\hline
\multirow{3}{*}{\textbf{Enron RCNN}}
&TextBugger&	1-18:55:48	&\textbf{42.42\%}&	38.68\%&	62.52\%&	61.51\%&	13.39\%&	17.31\%\\
	&TextFooler&	2-03:07:54&	20.98\%&	8.12\%&	61.19\%&	77.47\%&	7.23\%&	10.96\%\\
	&\textbf{ReinforceBug}	&\textbf{1-08:15:21}&31.80\%&10.37\%&\textbf{66.34\%}	&\textbf{88.17\%}	&\textbf{6.86\%}&\textbf{7.95\%}\\

	\hline
\multirow{3}{*}{\textbf{Twitter RCNN}}
&	TextBugger&	0-1:30:23&	\textbf{64.71\%}&	20.52\%&	75.6\%&	58.5\%&	26.43\%&	58.17\%\\
	&TextFooler	&0-1:34:05	&12.59\%&	\textbf{9.17\%}&	79.18\%&	74.65\%&	19.71\%&	14.01\%\\
	&\textbf{ReinforceBug}&	\textbf{0-0:47:29}&12.68\%&	13.43\%&	\textbf{81.77\%}&	\textbf{75.84\%}	&\textbf{16.97\%}&	20.17\%\\
	\hline
\multirow{3}{*}{\textbf{Yelp BiLSTM}}
& TextBugger&	2-13:26:43&	\textbf{50.42\%}&	26.73\%&	58.73\%&	66.96\%&	5.23\%&	8.21\%\\
	&TextFooler&	2-16:17:21	&39.33\%&	\textbf{4.00\%}&	58.85\%&	80.58\%&	3.15\%&	5.25\%\\
	&\textbf{ReinforceBug}	&\textbf{1-15:27:12}&46.19\%&5.77\%&\textbf{61.22\%}&	\textbf{89.07\%}&	3.62\%&\textbf{	4.42\%}\\
	\hline
\multirow{3}{*}{\textbf{Toxic BiLSTM}}
&	TextBugger&	0-11:57:28&	\textbf{33.99\%}&	36.24\%&	62.52\%&	61.49\%&	11.07\%&	18.65\%\\
	&TextFooler&	0-15:19:29	&16.05\%&	\textbf{7.93\%}&	65.83\%&	77.23\%&	8.37\%&	7.94\%\\
	&\textbf{ReinforceBug}	&\textbf{0-09:22:49}&30.16\%&10.52\%&\textbf{70.36\%}&\textbf{78.43\%}&\textbf{4.83\%}&	\textbf{4.13\%}\\
\hline

\end{tabular}}
\end{small}
\end{table*}
\begin{table*}[!htb]
\caption{Transferrability of AEs generated by ReinforceBug Attack on other models}
\label{attacktrans}
\centering
\begin{small}
\resizebox{\textwidth}{!}{\begin{tabular}{c|l|c|c|c|c|c|c|c|c|c|c|c}
\hline
	&\multicolumn{3}{c|}{\textbf{Enron}}		&\multicolumn{3}{c|}{\textbf{Twitter}}
	&\multicolumn{3}{c|}{\textbf{Yelp}}
	&\multicolumn{3}{c}{\textbf{Toxic}}	\\
	\cline{2-13}
	&\textbf{CharCNN}&	\textbf{WordCNN}&	\textbf{BILSTM}&	\textbf{CharCNN}	&\textbf{WordCNN}	&\textbf{BILSTM}	&\textbf{CharCNN}&\textbf{WordCNN}&\textbf{RCNN}&\textbf{CharCNN}&\textbf{WordCNN}&\textbf{RCNN}\\
	\hline
TextBugger	&\textbf{51.0\%}&	17.3\%&	\textbf{16.5\%}&	11.7\%&	30.8\%&	50.0\%&	21.8\%&	21.6\%&	24.5\%&	50.9\%&	43.8\%&	42.5\%\\
\hline
TextFooler&	47.0\%&	12.5\%&	12.3\%&	24.0\%&	41.6\%&	49.7\%&	25.6\%&	22.5\%&	27.1\%&	66.3\%&	59.8\%&	52.9\%\\
\hline
\textbf{ReinforceBug}&	47.3\%&	\textbf{27.5\%}&	16.3\%&	\textbf{53.5}\%&	\textbf{42.6\%}&	\textbf{52.0\%}&\textbf{34.0\%}&	\textbf{37.2\%}&	\textbf{34.5\%}&	\textbf{70.3\%}&\textbf{	64.9\%}&\textbf{	72.2\%}\\
\hline
\end{tabular}}
\end{small}
\end{table*}
\subsection{Knowledge Transferability}
\label{datasettransfer}
Table~\ref{testing} shows the transferability of policy learned by ReinforceBug on unseen datasets for each task, for benchmarking the results are also compared with state-of-the-art attacks TextBugger and Textfooler.
It is evident from the results that \emph{ReinforceBug} takes less time to generate adversarial samples as compared to the other models. 
It is because \emph{ReinforceBug} utilizes the same important word vocabulary selected while learning and has already explored and learned utility-preserving transformations on them during training.
Additionally, although TextFooler and TextBugger both perform example specific perturbation and \emph{ReinforceBug} might suffer from out-of-vocabulary words
but still the success rate of \emph{ReinforceBug} on test datasets is more than TextFooler attack and less than TextBugger which are aligned with our finding in section \ref{evaluate}. Also, from Table~\ref{testing} it is evident that our \emph{ReinforceBug} produces utility-preserving AEs with high PCS and semantic similarity (i.e., on average more than 70\% and 82\% respectively) and comparatively low perturbation rate (on average 8.5\%) for all test datasets.
It suggests that the important words and transformation learned from the training datasets are transferable to unseen datasets and can generalize the vulnerabilities of the target models.
\begin{figure*}[!t]
\begin{subfigure}{.23\textwidth}
  \centering
  \includegraphics[width=.99\linewidth]{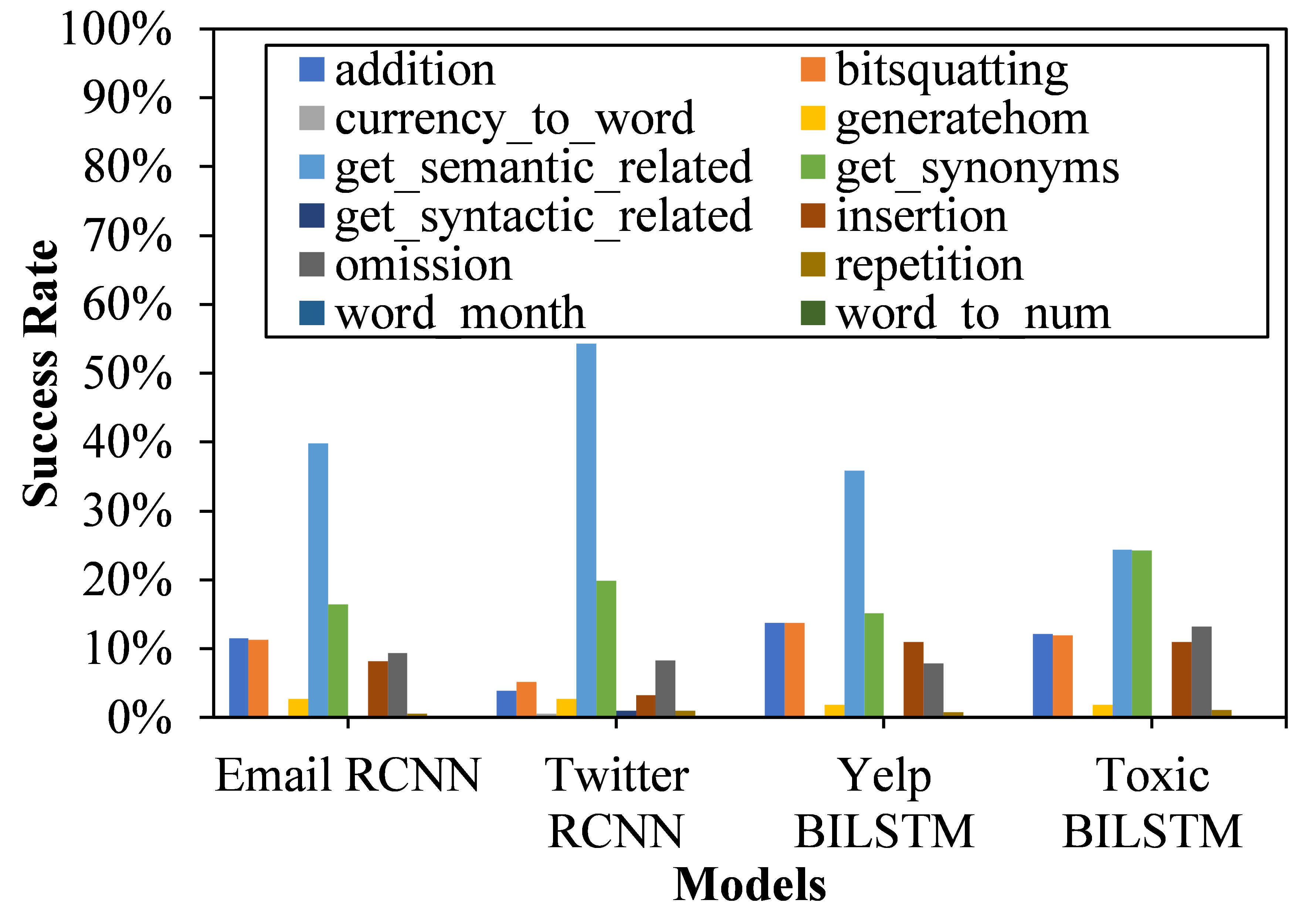}
\caption{ Action Type Distribution across Trained Models}
\label{fig:sub-first}
\end{subfigure}
\begin{subfigure}{0.75\textwidth}
\begin{small}
\centering
\includegraphics[width=.99\linewidth]{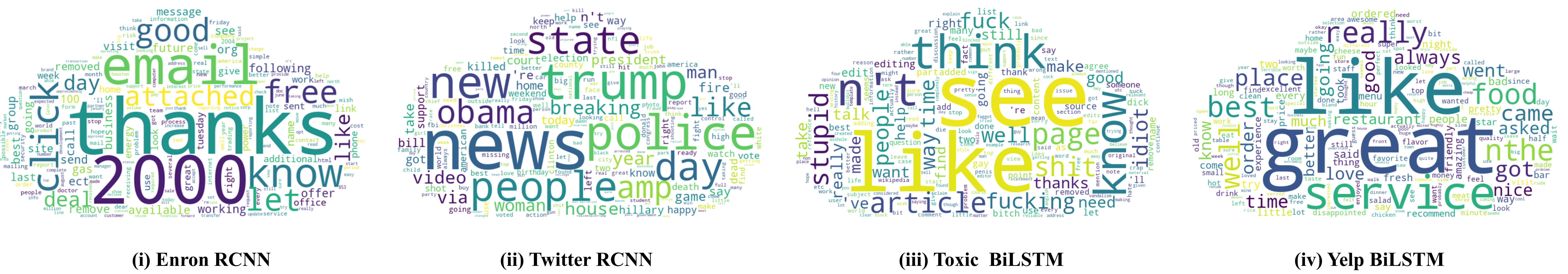}
\caption{Word Cloud of important words versus the number of Examples affected for the target models}
\label{fig:sub-second}
\end{small}
\end{subfigure}
\caption{Analysis of AEs generated by ReinforceBug}
\label{fig:fig}
\end{figure*}
\subsection{Attack Transferability on Models}
\label{modeltransfer}
To determine whether AEs curated based on one model can also fool other models for the same task, we examined the transferability of AEs on other models (see Table~\ref{tab2} for the performance of these models).
Table \ref{attacktrans} shows the transferability result. 
AEs generated by our method are more transferable to other models in comparison to the state-of-the-art attacks. 
However, there is a moderate degree of transferability between models, and the transferability is higher in the Twitter and Toxic detection task than in the Email and Yelp classification task. Nevertheless, BiLSTM trained on Enron dataset (having 97.60\% accuracy) offers more resilience to AEs generated by RCNN by limiting the success rate of the attack to (<17\%) for all the attacks, while
other models are highly vulnerable to the AEs.
It signifies that vulnerabilities exploited by our AEs are task-specific and moderately model-independent. 
\section{Analysis and Discussion}
\label{discussion}
\subsection{Action Type Distribution}
Figure~\ref{fig:sub-first} illustrates the proportion of each action type chosen by the ReinforceBug to generate utility-preserving AEs.
We can see that get\_semantic, and get\_synonyms action types are most dominant for all the tasks. One reason could be that get\_semantic is deliberately designed for creating similar contextual adversarial texts without deleting the original important word, while get\_synonyms replaces the word with similar meaning word. That is why the semantic similarity remains intact without impacting the linguistic structure of the text.
Other action types, i.e., addition, insertion, bitsquatting and omission that cause common typos are moderately chosen by our method, however, generatehom is rarely selected by \emph{ReinforceBug}, this is because it produces fewer replacement options for a word than other action types as it can only replace a character with a visually similar character. 
This reason is also valid of word\_month and word\_to\_num as only few words are either words representing month or numbers in the corpus vocabulary.

\subsection{Distribution of words versus the number of examples effected}
To demonstrate the knowledge transferability, we visualize the identified important words
according to the number of examples affected by their replacements in Figure~\ref{fig:sub-second}.
Here, the words impacting more examples are represented with a larger font.
Figure~\ref{fig:sub-second}(i) shows that in emails words such as click, attached, thanks and deal are more likely to affect the prediction of the target models by decreasing the spam intent to benign.
Whereas for twitter dataset, Figure~\ref{fig:sub-second}(ii) shows that the targeted RCNN model is more vulnerable to minor perturbation on words such as news, trump, obama, state and police.
For toxic content detection, the model decision is manipulated for most of the examples with words like see, like, think. 
Perturbing words like stupid and idiot decreases the toxicity of the text.
Lastly, for yelp dataset, changing words such as like, great and best increases the negative extent of the text. 
Therefore, it is evident that models are vulnerable to these words irrespective of a specific example; instead, these vulnerabilities affect multiple examples in the corpus and are transferable to new datasets as seen in (section \ref{datasettransfer}) as well as are transferable to other models (section \ref{modeltransfer}).
\section{Conclusion}
\label{conclusion}
Overall, this study proposes \emph{ReinforceBug}, a reinforcement learning-based framework to generate utility-preserving AEs against state-of-the-art text classifiers under black-box settings. 
Extensive experiments demonstrate that ReinforceBug
is effective in generating utility-preserving AEs that are transferable to other models and the learned policy is transferable to the unseen datasets.
\emph{ReinforceBug} identifies semantic concatenation and synonym substitution attacks as a significant threat to text-classifiers and suggests defence against these attacks should be explored in future to improve their robustness.
\section{Acknowledgements}
This work was supported with super-computing resources provided by the Phoenix HPC service at the University of Adelaide.
\bibliography{anthology,custom}
\bibliographystyle{acl_natbib}
\end{document}